\documentclass[conference]{IEEEtran}
\IEEEoverridecommandlockouts
\usepackage[T1]{fontenc}
\usepackage{cite}
\usepackage{amsmath,amssymb,amsfonts}
\usepackage{algorithmic}
\usepackage{graphicx}
\usepackage{textcomp}
\usepackage{xcolor}
\usepackage{wrapfig}
\usepackage{footmisc}
\usepackage{amsmath}
\def\BibTeX{{\rm B\kern-.05em{\sc i\kern-.025em b}\kern-.08em
    T\kern-.1667em\lower.7ex\hbox{E}\kern-.125emX}}

\newcommand{\equalcontribution}{\textsuperscript{$*$}}
\newcommand{\correspondingauthor}{\textsuperscript{$^\dagger$}}

\begin{document}

\title{LocRef-Diffusion:Tuning-Free Layout and Appearance-Guided Generation
}
\author{
\IEEEauthorblockN{Fan Deng \equalcontribution}
\IEEEauthorblockA{\textit{MetaX} \\
Shanghai, China \\
fan.deng@metax-tech.com}
\and
\IEEEauthorblockN{Yaguang Wu \equalcontribution}
\IEEEauthorblockA{\textit{MetaX} \\
Shanghai, China \\
yaguang.wu@metax-tech.com}
\and
\IEEEauthorblockN{Xinyang Yu \correspondingauthor}
\IEEEauthorblockA{\textit{MetaX} \\
Shanghai, China \\
xinyang.yu@metax-tech.com}
\and
\IEEEauthorblockN{Xiangjun Huang}
\IEEEauthorblockA{\textit{MetaX} \\
Shanghai, China \\
xiangjun.huang@metax-tech.com}
\and
\IEEEauthorblockN{Jian Yang}
\IEEEauthorblockA{\textit{MetaX} \\
Shanghai, China \\
jian.yang@metax-tech.com}
\and
\IEEEauthorblockN{Guangyu Yan}
\IEEEauthorblockA{\textit{MetaX} \\
Shanghai, China \\
guangyu.yan@metax-tech.com}
\and
\IEEEauthorblockN{Qiang Xu}
\IEEEauthorblockA{\textit{MetaX} \\
Shanghai, China \\
qiang.xu@metax-tech.com}
}

\maketitle
\begin{abstract}
    Recently, text-to-image models based on diffusion have achieved remarkable success in generating high-quality images. 
    However, the challenge of personalized, controllable generation of instances within these images remains an area in need of further development.
    In this paper, we present LocRef-Diffusion, a novel, tuning-free model capable of personalized customization of multiple instances' appearance and position within an image.
    To enhance the precision of instance placement, we introduce a Layout-net, which controls instance generation locations by leveraging both explicit instance layout information and an instance region cross-attention module. 
    To improve the appearance fidelity to reference images, we employ an appearance-net that extracts instance appearance features and integrates them into the diffusion model through cross-attention mechanisms.
    We conducted extensive experiments on the COCO and OpenImages datasets, and the results demonstrate that our proposed method achieves state-of-the-art performance in layout and appearance guided generation.
\end{abstract}

\begin{IEEEkeywords}
    text-to-image, tuning-free, personalization.
\end{IEEEkeywords}

\footnotetext[\value{footnote}]{\equalcontribution Equal contribution. \correspondingauthor Corresponding author.}
\begin{figure*}[htp]
\centering
\includegraphics[width=1.55\columnwidth]{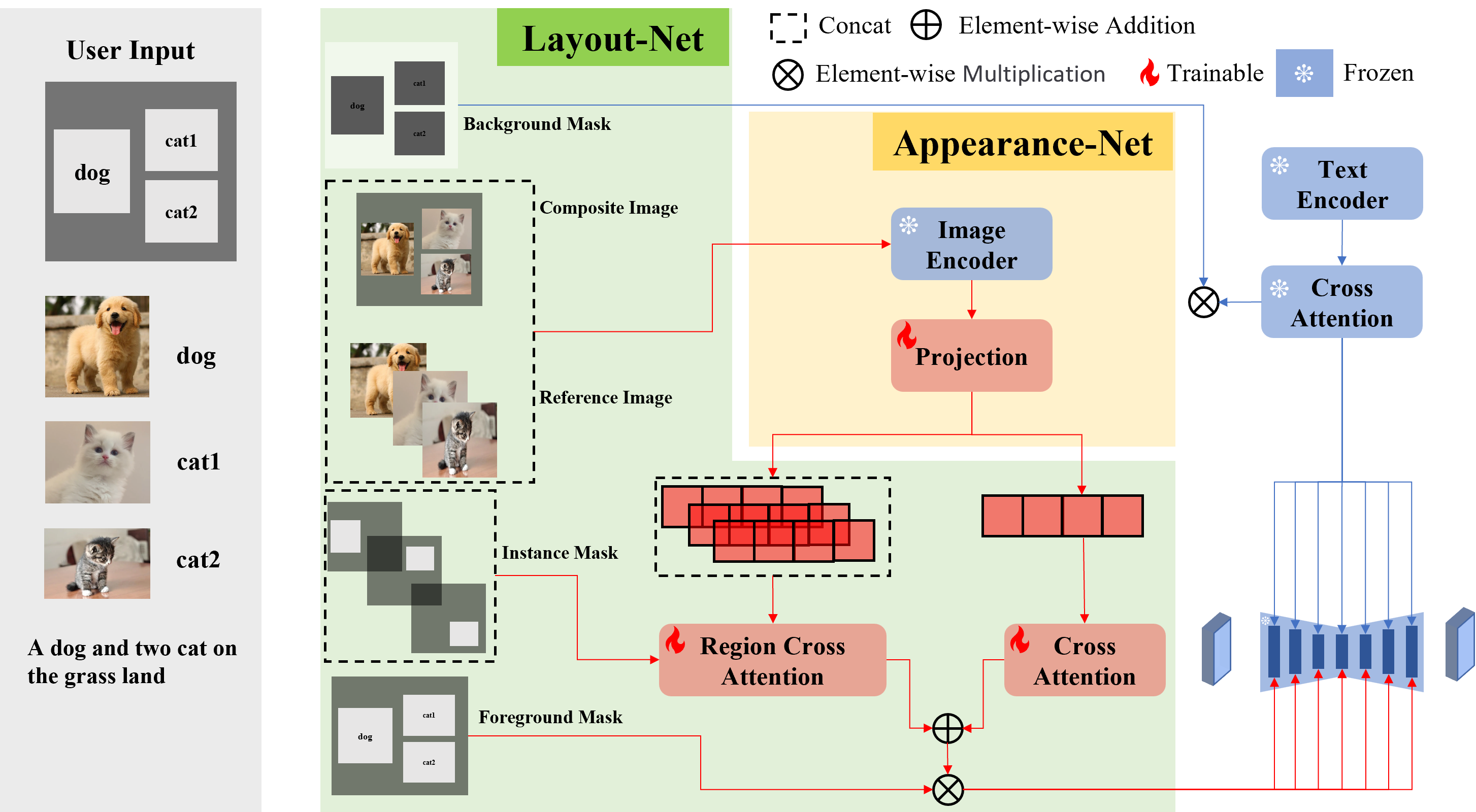}
\caption{The overall architecture of our proposed LocRef-Diffusion with Layout-net and Appearance-Net. Only the newly
added modules (in red color) are trained while the pretrained text-to-image model is frozen.}
\label{fig:your_label}
\end{figure*}
\section{Introduction}
Recent advancements in image generation models, particularly those trained on large-scale datasets \cite{ref1}, have significantly improved the quality and diversity of generated images. 
Although text-based guidance is valuable, it may lack the precision and intuitiveness needed for controlling the output. 
To address this, conditioning methods such as edges, normal maps \cite{ref2}, and semantic layouts \cite{ref3,ref4,ref5,ref6} have been introduced to offer finer control. 
Some methods \cite{ref7,ref8,ref9,ref10,ref11,ref12} use reference images to guide appearance features. 
These methods can be categorized into fine-tuning and tuning-free approaches. Among tuning-free methods, adapter-based approaches \cite{ref9} have shown strong performance in both image quality and computational efficiency.
However, when generating images with multiple instances, existing methods often exhibit deficiencies in localization precision and reference image fidelity. 
Specifically, generated objects often fail to align accurately with designated locations.
With regard to fidelity, there is a low degree of similarity between the generated objects and reference images. 
Additionally, feature leakage becomes more pronounced as the number of target instances increases.

To address the aforementioned issues, we propose a layout and reference image-guided tuning-free image personalization framework. 
Firstly, we design an image projection module named Appearance-Net to extract the detailed features of instances. 
Appearance-Net can effectively extract foreground features of instances while suppressing most background noise, thereby achieving high-fidelity feature transfer. 
Secondly, Layout-net is designed to precisely control the placement of objects within the scene.
Region-aware cross-attention is applied to avoid cross-attention leakage among multiple instances. 
Our proposed method ultimately achieves fine-grained control over the position and appearance of multiple subjects in the image. 
Experimental results demonstrate that our method outperforms state-of-the-art techniques in layout and appearance guided generation.

Our contributions are summarized as follows:
\begin{itemize}
\item To advance the development of controllable vision generation, we introduce the LocRef-Diffusion model to precisely guide the positioning of multiple objects while ensuring high-fidelity feature transfer. 
\item By preserving the pre-trained weights of the base model and incorporating new projection modules and localization layers, our model enables tuning-free open-world customized image generation.
\item Extensive experiments on COCO and OpenImages demonstrate that our model's zero-shot performance on customized image generation tasks significantly outperforms previous state-of-the-art methods.
\end{itemize}

\section{Related Work}
\subsection{Text-to-Image Generation}\label{AA}
Recent advancements in generative models have highlighted the success of diffusion model \cite{ref13}.
DALL-E2 \cite{ref14} employs a diffusion prior module in conjunction with a series of diffusion decoders to create high-resolution images. 
Imagen eliminates the need for latent representations by directly applying diffusion to pixel-level data.
Latent diffusion models (LDMs) \cite{ref16} execute this diffusion process within the latent space of a variational autoencoder \cite{ref17}.
Among these advancements, the development of Stable Diffusion has been particularly rapid.
\subsection{Layout-to-Image Generation}\label{AB}
Because text lacks the ability to precisely determine the positioning of generated instances.
Some Layout-to-Image approaches build upon pre-trained T2I models by incorporating layout information into the generation process
GLIGEN\cite{ref7} incorporates additional gated self-attention layers into pre-trained diffusion models to manage layout control, 
while MIGC\cite{ref19} employs innovative layout attention modules specifically designed for handling bounding boxes.
\subsection{Customized Image generation}\label{AC}
Recently, there has been a growing interest in developing tuning-free methods for stylized image generation, 
as seen in \cite{ref9,ref11}. 
These methods utilize lightweight adapters to extract image features and inject them into the diffusion process via self-attention or cross-attention. 
IP-Adapter\cite{ref9} and InstantStyle\cite{ref11} share a similar concept where a decoupled cross-attention mechanism is introduced to separate the cross-attention layers for text features and image features. 
StyleAlign\cite{ref20} swap the key and value features of the self-attention block in the original denoising process with those from a reference denoising process. 
Moreover, although these methods can generate high-fidelity images, users cannot specify the scenario and location of the target object.
\begin{figure}[htp]
\centering
\includegraphics[width=\columnwidth]{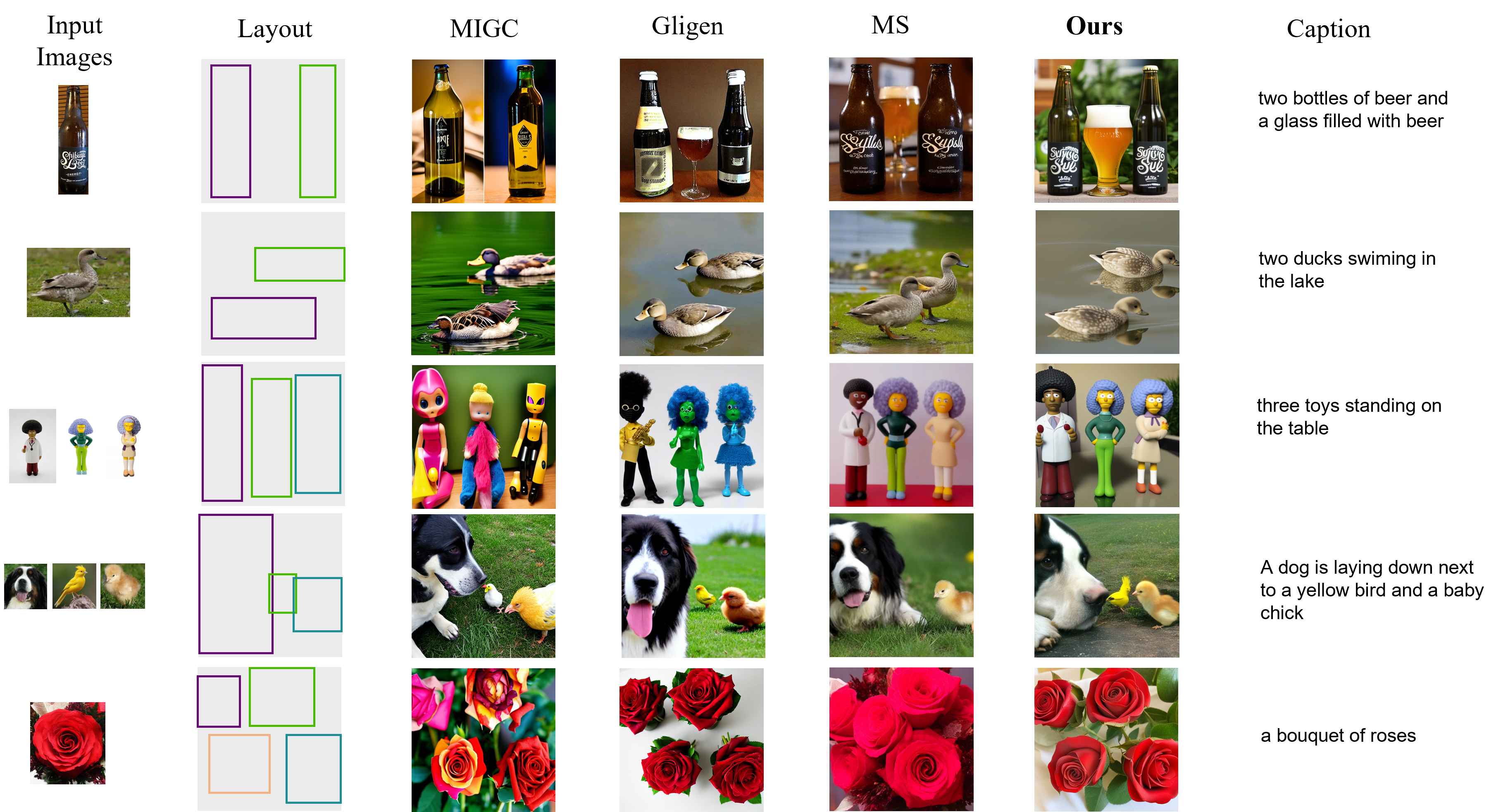} 
\caption{Qualitative result for multi-instance generation.}
\label{fig:your_label}
\end{figure}
\section{Method}
The pipeline of LocRef-Diffusion is illustrated in Fig.1. Given target objects, their specified locations, and a caption, LocRef-Diffusion generates a cohesive object-scene composition with high fidelity and diversity.
To accomplish these outcomes, we introduce two key components: Layout-net and Appearance-Net. Layout-net is designed to precisely control the placement of objects within the scene, ensuring that they are positioned according to the specified locations. This is achieved by leveraging explicit layout information and an instance region cross-attention module, which guides the diffusion process to generate objects in the correct locations.
Appearance-Net, on the other hand, is responsible for extracting appearance features from reference images. By integrating these features into the diffusion process through cross-attention mechanisms, Appearance-Net ensures that the generated objects maintain a high degree of similarity to the reference images, thereby preserving the fidelity and realism of the generated content.
Collectively, these components empower LocRef-Diffusion to generate highly realistic and customizable object-scene compositions that accurately reflect the input specifications, while maintaining diversity and visual quality at the forefront of current text-to-image generation models.

\subsection{Observations}\label{BA}
\textbf{Imprecise localization.}
Previous approaches, such as GLIGEN, MIGC, LayoutDiffusion\cite{ref21}, typically convert the coordinates into embeddings through Fourier transforms. 
GeoDiffusion\cite{ref22} discretizes the continuous coordinates by dividing the image into a grid of location bins, and each location bin is converted into location embedding. 
These embeddings are then integrated with the corresponding text features and embedded into the diffusion process. 
We have observed that Fourier transforms and grid discretization methods can introduce inaccuracies in positional control, often resulting in objects that are either excessively large or insufficiently sized relative to the designated bounding volume.
These limitations highlight the need for more precise methods to accurately represent spatial coordinates within designated boundaries.

\textbf{Cross-Attention Leakage.}
As discussed in Previous works\cite{ref19,ref21}, 
When addressing multi-instance prompts, Stable Diffusion faces challenges with attribute leakage, which can manifest in two primary forms:
Textual Leakage, owing to the causal attention masks utilized in the CLIP encoder, the tokens representing subsequent instances may exhibit semantic confusion. 
This phenomenon arises when the encoder fails to adequately distinguish between the attributes of different instances.
Spatial Leakage, the cross-attention mechanism in Stable Diffusion lacks precise position control\cite{ref19}, resulting in instances affecting the generation of each other's regions. 
Addressing these forms of leakage is crucial for achieving high-fidelity and semantically coherent multi-instance generation.
\subsection{Appearance-Net}\label{BB}
We employ a pre-trained CLIP image encoder to extract image features from visual prompts. 
During the training phase, the CLIP image encoder remains frozen. 
Given that reference images typically consist of both foreground and background elements, our objective is to effectively extract foreground features. 
While an instance segmentation model could be utilized to isolate a cleaner foreground, this approach introduces additional computational overhead. 
Moreover, our experiments will demonstrate that even without incorporating an instance segmentation model, 
our appearance network can effectively extract relatively clean foreground features. 
To achieve this, we employ a trainable projection network designed to differentiate between foreground and background features. 
Once trained to convergence, this projection network predominantly outputs foreground features while suppressing most background noise. 
The projection network used in this study comprises two linear layers and a normalization layer.
\subsection{Layout-net}\label{BC}
In contrast to previous approaches\cite{ref7,ref19,ref22}, our approach involves the following steps to precisely control the generation boundaries of the foreground and simultaneously avoid cross-attention leakage as shown in Fig. 1.
\begin{itemize}
\item First, on a 512x512 black background image, we designate multiple bounding boxes as the regions where the foreground will be generated.
Each corresponding foreground reference image is resized and placed into its respective region within these bboxes, creating a composite image (CI). Subsequently, the CI undergoes feature extraction via the Appearance-Net, resulting in the CI-embedding.
\item Second, for each reference image (RI), feature extraction is performed independently through the Appearance-Net, producing RI-embeddings.
\item We add a new trainable cross-attention layer to each cross-attention layer in the original UNet model to insert the CI-embedding, thereby obtaining the hidden states corresponding to the CI-embedding (CIHS).
\item Fourth, we add another new trainable region cross-attention layer to insert the RI-embeddings, thereby obtaining the hidden states corresponding to the RI-embeddings (RIHS). Here, for each instance, to avoid cross-attention leakage, we apply a instance-mask corresponding to the bbox to restrict the attention region, ensuring that the RI embedding influences only the hidden states within the bbox region.
\item Fifth, We generate both the foreground mask (FM) and the background mask (BM) from the bounding boxes. 
\end{itemize}
Then, the final hidden states (HS) of the cross-attention layer can be carried out using the following method:
\begin{equation}
    \scalebox{0.8}{$
    HS = HS*BM + (CIHS + RIHS)*FM
    $}
\end{equation}
\section{Experiments}
\subsection{Implementation Details}\label{CA}
\textbf{Data.} To train the LocRef-Diffusion, we build a dataset from the COCO2017 and OpenImage. BLIP is used to generate consistent image descriptions.

\textbf{Details.} LocRef-Diffusion is built upon the pre-trained sd1.5\cite{ref16} model. 
During training, the original parameters of base model remain frozen, while only the Appearance-Net and Layout-net are trained. 
The total trainable parameters of Appearance-Net and Layout-net is about 56M, making the LocRef-Diffusion quite lightweight.
We use AdamW with a learning rate set to 0.0001 and a weight decay of 0.01, and train the model for 100 epochs with batch size 32. 
During training, we resize the shortest side of the image to 512 and then center crop the image with 512x512 resolution.
We employed several data augmentation to enhance the diversity of the training data, such as random horizontal flips and rotations within 10 degrees.
Furthermore, the loss function includes enhanced weighting for foreground regions, with $W_{f}$ set to 2.3 and $W_{b}$ set to 0.3, guiding the model to focus on fidelity to the reference images.
For inference, we use the DDIM\cite{ref23} scheduler with 50 iteration steps.
\begin{equation}
    \scalebox{1}{$
    loss = loss*(BM*W_{b}+FM*W_{f})
    $}
\end{equation}
$W_{f}$ is foreground region loss weight, $W_{b}$ is background region loss weight.

\subsection{Evaluation Metrics}\label{DA}
\textbf{Localization Assessment.} The accuracy of the generated object positions is evaluated using YOLO detection scores. 
Specifically, we utilize a pre-trained YOLOv8 model to detect the bounding boxes of the generated objects and compare them with the input reference boxes. 
The precision of the model in generate target localization is quantified by the mean Average Precision (mAP) metric.

\textbf{Visual Assessment.}
We measure the similarity between the generated object and the reference object using the CLIP-I similarity score. 
This involves calculating the CLIP visual similarity between the generated target and the reference target. 
The similarity score ranges from 0 to 1, with values closer to 1 indicating a higher proficiency in shape transfer.
\begin{table}[!ht]
\caption{}\label{tab:tablenotes}
\begin{center}
\includegraphics[width=0.9\columnwidth]{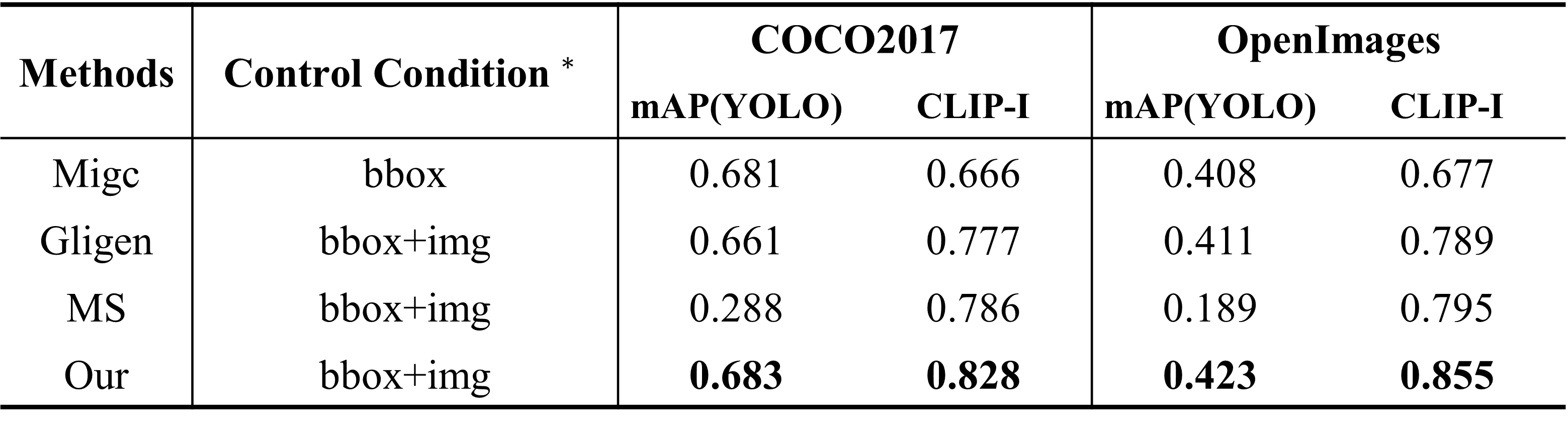}
\end{center}
\begin{center}
\parbox[t]{0.45\textwidth}{
    \textbf{Quantitative comparison between LocRef-Diffusion and baselines on COCO2017 and openImage.} The 'Control Conditions' column represents various conditions for controlling image generation, bbox represents the location of the generated target, and img represents the image referenced by the generated target. \\
}
\end{center}
\end{table}
\vspace{-1em}
\subsection{Results}\label{EA}

We compared our method with other SOTA approaches, including GLIGEN, MS-Diffusion\cite{ref24}, and MIGC. 
All three methods support multi-instance localization, with MIGC focusing solely on localization accuracy, while GLIGEN and MS-Diffusion additionally offer personalized customization capabilities based on reference images.
The comparative results on COCO2017 and OpenImages are presented in Table I.
Our method exhibits superior performance in appearance fidelity, as evidenced by the CLIP-I metric, and it significantly outperforms other models.
Furthermore, even when compared to state-of-the-art models that only include target localization functionality, our method also achieves strong performance in target localization.
The qualitative results in Fig. 2 still show that our method outperforms other SOTA methods. Specifically, GLIGEN exhibits significant feature leakage, MS results in poorer localization.

\subsection{Ablation Study}\label{FA}

To validate the effectiveness of our proposed method, we conducted a series of ablation experiments focusing on three key modules of the Layout-net: Generated Region Guidance, Instance-Aware Fusion, and Instance Segmentation. 
As a contrast, the base model represents using only the composite image (CI) mentioned in Layout-net to guide image generation.    
The results, summarized in Table II, demonstrate that all modules contribute positively to the final outcome.

\textbf{Generation Region Guidance.} As shown in the table, incorporating Generation Region Guidance significantly improves localization performance, with the YOLOv8 mAP increasing from 0.232 to 0.548 (comparison between configurations 1 and 2). 
This indicates that our proposed Generation Area Guidance module substantially enhances target localization accuracy.

\textbf{Instance Aware Fusion.} The addition of the Instance Aware Fusion module further boosts the injection of instance-specific features, such as class, coordinates, and image information, into the generated targets. 
This leads to improvements in both localization and visual scores, with a particularly notable enhancement in visual performance (comparison between configurations 2 and 3).

\textbf{Instance Segmentation.}
Although the Appearance-Net can already effectively extract foreground object features and filter out most background features, a comparison of configurations 3 and 4 clearly demonstrates that providing an explicit foreground segmentation enhances localization accuracy. 
This improvement is primarily attributed to the fact that, within the test set, some primary target objects possess appendages. 
Without the application of a foreground mask filter, the model tends to include these appendages in the generated region, thereby disrupting the layout and leading to an uncontrolled generation range for the main target. 
However, this result also indicates that our Appearance-Net can effectively extract features of the foreground targets, regardless of whether additional instance segmentation information is provided.
\begin{table}[!ht]
\begin{center}
\caption{}\label{tab:tablenotes}
\begin{tabular}{|c|c c c|c|c|}
\hline
\textbf{Index} & \textbf{GRG} & \textbf{IAF} & \textbf{IS} & \textbf{YOLO score (mAP)}&\textbf{CLIP-I}\\
\hline
1 & & &  & 0.232 & 0.799 \\
\hline
2 & + & & & 0.548 & 0.808 \\
\hline
3 & + & + & & 0.62 & 0.812 \\
\hline
4 & + & + & + & 0.683 & 0.828 \\
\hline
\end{tabular}
\end{center}
\begin{center}
\parbox[t]{0.45\textwidth}{
    \textbf{Ablation study of LocRef-Diffusion.} Index 1 indicates that these three modules are not added for the base model for ablation comparison. Ablation experiments include Generation Region Guidance(\textbf{GRG}), Instance Aware Fusion(\textbf{IAF}), Instance Segmentation(\textbf{IS}). \\
}
\end{center}
\end{table}
\vspace{-1em}
\section{Conclusions}
In this work, we introduced LocRef-Diffusion, a novel, tuning-free model capable of personalized customization of multiple instances' appearance and position within an image.
The Appearance-Net is used to extract detailed features of instances, thereby achieving high-fidelity foreground instances feature transfer.
The Layout-net is designed to precisely control the placement of instances within the image, ensuring that they are positioned according to the specified locations. 
Comprehensive experiments are conducted on COCO2017 and OpenImages.
Experiment results verify the localization accuracy and appearance consistency of our LocRef-Diffusion.

\end{document}